\documentclass[11pt,a4paper]{article}
\usepackage{authblk} 
\usepackage{times}
\usepackage{url}
\usepackage{latexsym}

\usepackage{booktabs} 
\usepackage{multirow} 
\usepackage{calc}	  
\usepackage{tikz}     
\usetikzlibrary{calc} 
\usepackage[colorlinks=true,allcolors=blue!60!black,breaklinks]{hyperref} 
\usepackage{eacl2017} 

\eaclfinalcopy 
\def\eaclpaperid{361} 

\newcommand{\sym}[1]{\ensuremath{\mathtt{#1}}} 
\newcommand{\bs}{\backslash} 
\newcommand{\NP}{N\!P} 
\newcommand{\tokglue}{\textvisiblespace} 
\newcommand{\forapp}{\scalebox{.8}{$>$}} 
\newcommand{\backapp}{\scalebox{.8}{$<$}} 
\newcommand{\lam}[1]{\ensuremath{\lambda#1.}} 
\newcommand{\drsvar}{D\raisebox{-1.5mm}{\kern-1.3ex\rotatebox{90}{=}}\kern.7ex}
\newcommand{\toppad}[1]{\raisebox{#1}{}}

\newcommand{\lexitem}[5]{%
\textsc{#1}\\[-1mm]
\sym{#2}\\[-1mm]
\small#3\\[-1mm]
\scriptsize#4\\[-1mm]
\scriptsize#5%
}

\newcommand{\CCG}[3][]{%
\begin{tabular}[t]{@{}c@{}}
\begin{tabular}[t]{@{}c@{}}#3\end{tabular}
\\[-3mm]
\rule[2.3pt]{\maxof{\widthof{#3}}{\widthof{\begin{tabular}{@{}c@{}}#2\end{tabular}}}*\real{0.9}-\widthof{#1}}{.5pt}#1
\\[-2.5mm]
#2
\end{tabular}}

\newcommand{\drs}[2]
{{\begin{tabular}{@{}c@{}}
\noalign{\smallskip}
\begin{tabular}{|@{\kern3pt}l@{\kern2pt}|}\hline
#1\toppad{2ex}\\\hline#2\\\hline
\end{tabular}
\smallskip
\end{tabular}}%
}

\newcommand{\parallelmeaningbank}{Parallel Meaning Bank}
\newcommand{\pmb}{PMB}

\title{The \parallelmeaningbank: \ Towards a Multilingual Corpus of\\Translations Annotated with Compositional Meaning Representations}

\author[1]{\bf Lasha Abzianidze}
\author[1]{\bf Johannes Bjerva}
\author[1]{\bf Kilian Evang}
\author[1]{\bf Hessel Haagsma}
\author[1]{\\ \bf Rik van Noord}
\author[2]{\bf Pierre Ludmann}
\author[3]{\bf Duc-Duy Nguyen}
\author[1]{\bf Johan Bos}
\affil[1]{CLCG, University of Groningen, The Netherlands}
\affil[2]{\'Ecole Normale Sup\'erieure de Cachan, France}
\affil[3]{University of Trento, Italy}
\affil[ ]{\tt \{l.abzianidze,j.bjerva,k.evang\}@rug.nl}
\affil[ ]{\tt \{hessel.haagsma,r.i.k.van.noord,johan.bos\}@rug.nl}
\affil[ ]{\tt pierre.ludmann@ens-cachan.fr}
\affil[ ]{\tt ducduy.nguyen@studenti.unitn.it}

\date{}

\begin{document}
\maketitle
\begin{abstract}
The \parallelmeaningbank\ is a corpus of translations annotated with
shared, formal meaning representations comprising over 11 million words divided over four languages (English, German, Italian, and Dutch). Our approach is based on cross-lingual projection: automatically produced (and manually corrected) semantic annotations for English sentences are mapped onto their word-aligned translations, assuming that the translations are meaning-preserving.
The semantic annotation consists of five main steps: (i) segmentation of the text in sentences and lexical items; (ii) syntactic parsing with Combinatory Categorial Grammar; (iii) universal semantic tagging; (iv) symbolization; and (v) compositional semantic analysis based on Discourse Representation Theory. These steps are performed using statistical models trained in a semi-supervised manner. The employed annotation models are all language-neutral. Our first results are promising.
\end{abstract}

\section{Introduction} 
There is no reason to believe that the ingredients of a meaning representation for one language should be different from that for another language. Hence, a meaning-preserving translation from a sentence to another language should, arguably, have equivalent meaning representations. Hence, given a parallel corpus with at least one language for which one can automatically generate meaning representations with sufficient accuracy, indirectly one also produces meaning representations for aligned sentences in other languages. The aim of this paper is to present a method that implements this idea in practice, by building a parallel corpus with shared formal meaning representations, that is, the \parallelmeaningbank\ (\pmb).

Recently, several semantic resources---corpora of texts annotated with meanings---have been developed to stimulate and evaluate semantic parsing. Usually, such resources are manually or semi-automatically created, and this process is expensive since it requires training of and annotation by human annotators. The AMR banks of Abstract Meaning Representations for English \cite{amr:13} or Chinese and Czech \cite{lrec:2014} sentences, for instance, are the result of manual annotation efforts. Another example is the development of the Groningen Meaning Bank \cite{GMB:2017}, a corpus of English texts annotated with formal, compositional meaning representations, which took advantage of existing semantic parsing tools, combining them with human corrections.

In this paper we propose a method for producing meaning banks for several languages (English, Dutch, German and Italian), by taking advantage of translations. On the conceptual level we follow the approach of the Groningen Meaning Bank project \cite{gmb:eacl}, and use some of the tools developed in it. 
The main reason for this choice is that we are not only interested in the final meaning of a sentence, but also in how it is derived---the compositional semantics. These derivations, based on Combinatory Categorial Grammar (CCG, Steedman, 2001\nocite{Steedman:01}), give us the means to project semantic information from one sentence to its translated counterpart.

\begin{figure*}[th!]
  \centering
  \includegraphics[keepaspectratio, width=1.0\textwidth]{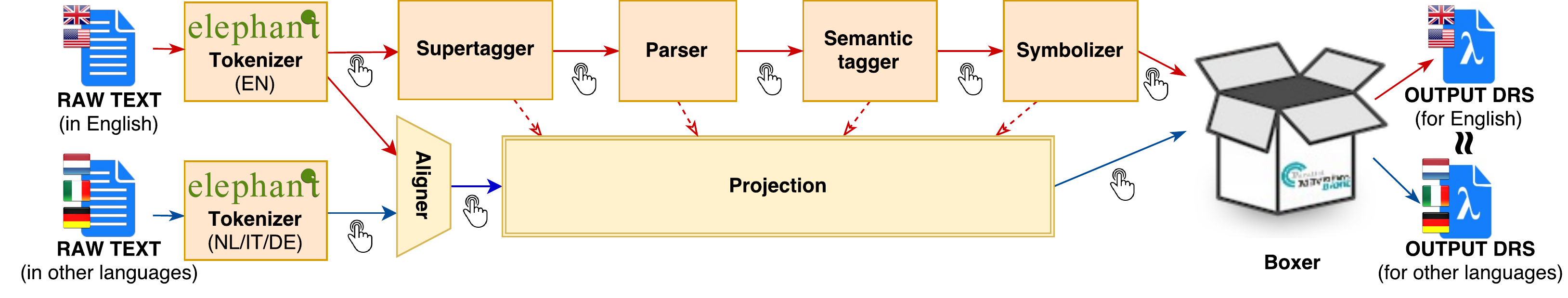}
  \vspace{-2mm}
  \caption{Annotation pipeline of the \pmb. Manual corrections can be added at each annotation layer.}
  \label{fig:PMBstructure}
\end{figure*}

The goal of the \pmb{} is threefold. First, it
will serve as a test bed for cross-lingual compositional semantics, enabling
systematic studies of the challenges arising from
loose translations and different semantic granularities.
The second goal is to produce data for building semantic
parsers for languages other than English. This, in turn,
will help with the third, long-term goal, which concerns the process
of translation itself. Human translators purposely change
meaning in translation to yield better
translations \cite{langeveld}. The third goal is thus to develop methods to automatically detect such shifts in meaning.

\section{Languages and Corpora}
The foundation of the \pmb{} is a large set of raw, parallel texts. Ideally, each text has a parallel version in every language of the meaning bank, but in practice, having a version for the pivot language (here: English) and one other language is sufficient for our purposes. 
Another criterion for selection is that freely distributable texts are preferable over texts which are under copyright and require (paid) licensing.

Besides English we chose two other Germanic languages, Dutch and German, because they are similar to English. We also include one Romance language, Italian, in order to test whether our method works for languages which are typologically more different from English.

The texts in the \pmb{} are sourced from twelve different corpora from a wide range of genres, including, among others: Tatoeba%
\footnote{\url{https://tatoeba.org}},
News-Commentary (via OPUS, Tiedemann, 2012\nocite{opus}), Recognizing Textual Entailment \cite{rte3}, Sherlock Holmes stories%
\footnote{\url{http://gutenberg.org}, \url{http://etc.usf.edu/lit2go}, ~~ \url{http://gutenberg.spiegel.de}}, 
and the Bible \cite{Christodouloupoulos2015}.

These corpora are divided over 100 parts in a balanced way. Initially, two of these parts, 00 and 10, are selected to be the gold standard (and thus will be manually annotated). This ensures that the gold standard represents the full range of genres.

The resulting corpus contains over 11.3 million tokens, divided into 285,154 documents. All of them have an English version. 72\% have a German version, 14\% a Dutch one and 42\% an Italian one. 9\% have German and Dutch, 6\% have Dutch and Italian and 18\% have Italian and German. 5\% exist in all four languages.

\section{Automatic Annotation Pipeline}
Our goal is first to richly annotate the English corpus, with annotations ranging from segmentation to deep semantics, and then project these annotations to the other languages via alignment.
The annotation consists of several layers, each of which will be presented in detail below.
Figure~\ref{fig:PMBstructure} gives an overview of the pipeline while Figure~\ref{fig:PMBexample} shows the annotation example.  

\subsection{Segmentation}
Text segmentation involves word and sentence boundary detection.
Multiword expressions that represent constituents are treated as single tokens.
Closed compound words that have a semantically transparent structure are decomposed.
For example, {\em impossible} is decomposed into {\em im} and {\em possible} while
{\em Las Vegas} and {\em 2 pm} are analysed as a single token.
In this way we aim to assign `atomic' meanings to tokens and avoid redundant lexical semantics.
Segmentation follows an IOB-annotation scheme on the level of characters, with four labels: beginning of sentence, beginning of word, inside a word, and outside a word. We use the same statistical tokenizer, Elephant \cite{elephant}, for all four languages, but with language-specific models.

\begin{figure*}[t] 
\begin{tikzpicture}
\tikzset{every node/.style={inner sep=2pt, outer sep=0pt, align=center}}
\tikzset{every edge/.style={line width=1pt,->,>=latex,draw=gray}}
\node(base)at(0,0){};
\node(he)at($(base.east)+(10mm,0mm)$){};
\node(came)at($(he.east)+(21mm,0mm)$){};
\node(back)at($(came.east)+(25mm,0mm)$){};
\node(at)at($(back.east)+(33mm,0mm)$){};
\node(clock)at($(at.east)+(50mm,0mm)$){};
\node(er)at($(he.north)+(0mm,10mm)$){\small $\NP$\\[-1mm]Er};
\node(kam)at($(came.north)+(0mm,10mm)$){\small $S\bs\NP$\\[-1mm]kam};
\node(um)at($(back.north)+(0mm,10mm)$){\small $((S\bs\NP)\!\bs\!(S\bs\NP))\!/\!\NP$\\[-1mm]um}; 
\node(uhr)at($(at.north)+(8mm,9.4mm)$){\small $N$\\[-1mm]f{\"u}nf\tokglue{}Uhr};
\node(zuruck)at($(clock.north)+(-10mm,10mm)$){\small $(S\bs\NP)\!\bs\!(S\bs\NP)$\\[-1mm]zur{\"u}ck};
\path (he.north) edge (er.south);
\path (came.north) edge (kam.south);
\path (at.north) edge (um.south);
\path (back.north) edge (zuruck.south);
\path (clock.north) edge (uhr.south);
\end{tikzpicture}
\vspace{-3mm}

\centerline{
\CCG[\backapp]{\small$S$}{
\CCG{\lexitem{pro}{male}{$\NP$}
		{$\lam{p}\drsvar*px$}
		{\drs{$x$}{\toppad{2.1ex}$\sym{male}(x)$}}}
	{He}~
\CCG[\backapp]{\small$S\bs\NP$}{
	\CCG[\backapp]{\small$S\bs\NP$}{
		\CCG{\lexitem{eps}{come}{$S\bs\NP$}
				{$\lam{G p}G (\lam{x}\drsvar; pe)$}
				{\drs{$e$ ~ $t_1$ ~ $t_2$}{\toppad{2.1ex}$\sym{come}(e)$\\$Theme(e,x)$\\$Time(e,t_2)$\\$now(t_1)$ \kern4mm $t_2 < t_1$}}}
			{came}
		\CCG{\lexitem{ist}{back}{$(S\bs\NP)\!\bs\!(S\bs\NP)$}
				{$\lam{VGp} VG(\lam{x} \drsvar; px)$}
				{\kern5ex\drs{$s$}{\toppad{2.1ex}$Manner(x,s)$\\$\sym{back}(s)$}}}
			{back}}~
	\CCG[\forapp]{\small$(S\bs\NP)\!\bs\!(S\bs\NP)$}{
		\CCG{\lexitem{rel}{at}{$((S\bs\NP)\!\bs\!(S\bs\NP))\!/\!\NP$}
				{$\lam{GV\!Hp} VH(\lam{x} G(\lam{y} \drsvar; px))$}
				{\kern19ex\drs{}{\toppad{2.1ex}$\sym{at}(x,y)$}}}
			{at}~
		\CCG[\forapp]{\small$\NP$}{
			\CCG{\lexitem{dis}{\emptyset}{$\NP\!/\!N$}
					{$\lam{pq} \drsvar; (px; qx)$}
					{\kern-4ex\drs{$x$}{}}}
				{$\emptyset$}
    		\CCG{\lexitem{clo}{17\!:\!00}{$N$}
    				{$\lam{x} \drsvar$}
    				{\drs{}{\toppad{2.1ex}$time(x)$\\$value(x,\sym{17\!:\!00})$}}}
    			{5\tokglue{}o'clock}}}}}
}\vspace{-2mm}

\centerline{\small
\drs{$x$~~ $y$~~ $e$~~ $s$~~ $t_1$~~ $t_2$}
{\begin{tabular}[t]{c c c c c}
$\sym{come}(e)$ & $Time(e,t_2)$ & 
$Theme(e,x)$ & $Manner(e,s)$ &
$\sym{at}(e,y)$ ~~ $time(y)$\toppad{2.1ex}\\
$now(t_1)$ & $t_2 < t_1$ &
$\sym{male}(x)$ & $\sym{back}(s)$ &
$value(y,\sym{17\!:\!00})$
\end{tabular}}}		
\caption{Document 00/3178: Projection of the annotation from English to German.
The source sentence is annotated, in this order, with semtags, symbols, CCG categories and lexical semantics.
The DRS for the whole sentence is obtained compositionally from the lexical DRSs.
}
\label{fig:PMBexample}			
\end{figure*}

\subsection{Syntactic Analysis}
We use CCG-based derivations for syntactic analysis. The transparent syntax-semantic interface of CCG makes the derivations suitable for wide-coverage compositional semantics \cite{Bosetal:04}. CCG is also a lexicalised theory of grammar, which makes cross-lingual projection of grammatical information from source to target sentence more convenient (see Section~\ref{sec:pro}). 

The version of CCG that we employ differs from standard CCG: in order to facilitate the cross-lingual projection process and retain compositionality, type-changing rules of a CCG parser are explicated by inserting (unprojected) empty elements which have their own semantics (see the token $\emptyset$ in Figure~\ref{fig:PMBexample}).

For parsing, we use EasyCCG \cite{lewisSteedman:14}, which was chosen because it is accurate, does not require part-of-speech annotation (which would require different annotation schemes for each language) and is easily adaptable to our modified grammar formalism.

\subsection{Universal Semantic Tagging}
To facilitate the organization of a wide-coverage semantic lexicon for cross-lingual semantic analyses, we develop a {\em universal semantic tagset}.
The semantic tags ({\em semtags}, for short) are language-neutral, generalise over part-of-speech and named entity classes, and also add more specific information when needed from a semantic perspective.
Given a CCG category of a token, we specify a general schema for its lexical semantics by tagging the token with a semtag. 

Currently the tagset comprises 80 different fine-grained semtags divided into 13 coarse-grained classes \cite{Bjervaetal:16}. 
We do not list all possible semtags here, but give some examples instead.
For instance, the semtag {\sc not} marks negation triggers, e.g., {\em not}, {\em no}, {\em without} and affixes, e.g., {\em im-} in \emph{impossible}; the semtag {\sc pos} is assigned to possibility modals, e.g., {\em might}, {\em perhaps} and {\em can}.
{\sc rol} identifies roles and professions, e.g., {\em boxer} and {\em semanticist}, while {\sc con} is for concepts like {\em table} and {\em wheel}.
Distinguishing roles from concepts is crucial to get accurate semantic behaviour.
\footnote{Roles are mostly consistent with each other while concepts are not. 
For instance, an entity can be a boxer and a semanticist at the same time but not a wheel and a table.}

We use the semantic tagger based on deep residual networks. It works directly on the words as input, and therefore requires no additional language-specific features.
The first results on semantic tagging, with an accuracy of 83.6\%, are reported by \newcite{Bjervaetal:16}.   

\subsection{Symbolization}
The meaning representations that we use contain logical symbols and non-logical symbols. The latter are based on the words mentioned in the input text. We refer to this process as \textit{symbolization}. It combines lemmatization with normalization, and performs some lexical disambiguation as well.
For example, \sym{male} is the symbol of the pronouns {\em he} and {\em himself}, \sym{europe} of the adjective {\em European}, and \sym{14\!:\!00} for the time expression {\em 2 pm}. 
A symbol together with a CCG category and a semtag are sufficient to determine the lexical semantics of a token (see Figure~\ref{fig:PMBexample}).
Some function words do not need symbols since their semantics are expressed with logical symbols, e.g., auxiliary verbs, conjunctions, and most determiners.

Notice that the employed symbols are not as radical and verbalized as the {\em concepts} in AMRs, e.g., the symbol of {\em opinion} is \sym{opinion} rather than \sym{opine}. 
First, using deep forms as symbols often makes it difficult to recover the original and semantically related forms, e.g., if {\em opinion} had the symbol \sym{opine}, then it would be difficult to recover {\em opinion} and its semantic relation with {\em idea}.
Second, alignment of translations does not always work well with deep forms, e.g., {\em opinion} can be translated as {\em parere} in Italian
and {\em mening} in Dutch, but it is unnatural to align their symbols to \sym{opine}.
After all, having such alignments would make it difficult to judge good and bad translations, which is one of the goals of the \pmb.

The symbolizer could either be implemented as a rule-based
system with multiple modules, or as a system that learns
the required transformations from examples. The advantage
of the latter is that it is more robust to typos
and other spelling variants without manual engineering.
To evaluate the feasibility of this approach,
we built a character-based sequence-to-sequence model with deep recurrent neural networks, which uses words,
semtags, and additional data from existing knowledge sources, such as WordNet \cite{wordnet}, Wikipedia, 
and UNECE codes for trade\footnote{\mbox{\url{http://unece.org/cefact/codesfortrade}}}, to do symbolization.
We are currently investigating how the performance of
machine learning-based symbolizer compares to a rule-based one incorporating the lemmatizer Morpha \cite{minnen:01}.

\subsection{Semantic Interpretation}
Discourse Representation Theory (DRT, Kamp and Reyle, 1993\nocite{DRT:93}), is the semantic formalism that is used as a semantic representation in the PMB.
It is a well-studied theory from a linguistic semantic viewpoint and suitable for compositional semantics.%
\footnote{In particular, we employ Projective DRT \cite{venhuizen2015PhDthesis}---an extension of DRT that accounts for presuppositions, anaphora and conventional implicatures in a generalized way.}
Expressions in DRT, called Discourse Representation Structures (DRSs), have a recursive structure and are usually depicted as boxes.
An upper part of a DRS contains a set of referents while the lower part lists a conjunction of atomic or compound conditions over these referents (see an example of a DRS in the bottom of Figure~\ref{fig:PMBexample}).

Boxer \cite{boxer}, a system that employs \mbox{$\lambda$-calculus} to construct DRSs in a compositional way, is used to derive meaning representations of the documents. 
However, the original version of Boxer is tailored to the English language. 
We have adapted Boxer to work with the universal semtags rather than English-specific part-of-speech tags.
Boxer also assigns VerbNet/LIRICS thematic roles \cite{Bonial:11} to verbs so that the lexical semantics of verbs include the corresponding thematic predicates (see {\em came} in Figure~\ref{fig:PMBexample}).

Hence an input to Boxer is a CCG derivation where all tokens are decorated with semtags and symbols.
This information is enough for Boxer to assign a lexical DRS to each token and produce a DRS for the entire sentence in a compositional and language-neutral way (see Figure~\ref{fig:PMBexample}).

\section{Cross-lingual Projection}
\label{sec:pro}
The initial annotation for Dutch, German and Italian is bootstrapped via word alignments. Each non-English text is automatically word-aligned with its English counterpart, and non-English words initially receive semtags, CCG categories and symbols based on those of their English counterparts (see Figure~\ref{fig:PMBexample}).

CCG slashes are flipped as needed, and 2:1 alignments are handled through functional composition. Then, the CCG derivations and DRSs can be obtained by applying CCG's combinatory rules in such a way that the same DRS as for the English sentence results \cite{evang2016,evang-thesis}.

If the alignment is incorrect, it can be corrected manually (see Section~\ref{sec:bow}).
The idea behind this way of bootstrapping is to exploit the advanced state-of-the-art of NLP for English, and to encourage parallelism between the syntactic and semantic analyses of different languages.

To facilitate cross-lingual projection, alignment has to be done at two levels: sentences and words.
Sentence alignment is initially done with a simple one-to-one heuristic, with each English sentence aligned to a non-English sentence in order, to be corrected manually. Subsequently, we automatically align words in the aligned sentences using \textsc{giza++} \cite{giza}.

Although we use existing tools for the initial annotation of English and projection as the initial annotation of non-English documents, our aim is to train new language-neutral models. Training new models on just the automatic annotation will not yield better performance than the combination of existing tools and projection. However, we improve these models constantly by adding manual corrections to the initial automatic annotation, and retraining them.
In addition, this approach lets us adapt to revisions of the annotation guidelines.

\section{Adding Bits of Wisdom}
\label{sec:bow}
For each annotation layer,  manual corrections can be applied to any of the four languages.
These annotations are called Bits of Wisdom (BoWs, following \newcite{gmb:eacl}), and they overrule the annotations of the models if they are in conflict.
Based on the BoWs, we distinguish three disjoint classes of annotation layers: gold standard (manually checked), silver standard (including at least one BoW) and bronze standard (no BoWs). Table~\ref{tab:layer_stats} shows how these classes are distributed across languages and documents.

\bgroup
\renewcommand{\arraystretch}{1}
\setlength{\tabcolsep}{6pt}
\begin{table}[htbp]
  \centering
  \begin{tabular}{lcrrr}
  \toprule
  \textbf{Layer} & \textbf{Lang} & \textbf{Gold} & \textbf{Silver} & \textbf{Bronze}\\ 
  \midrule
  \multirow{4}{*}{\textbf{Tokens}} 	  
    & \textsc{en}   & 6,810		& 2,548		& 275,796\\
    & \textsc{de}   & 4,757   	& 736		& 198,776\\
    & \textsc{it}   & 2,843   	& 384		& 117,792\\
    & \textsc{nl}   & 945    	& 528 		& 38,942\\
  \midrule
  \textbf{Semtags}      
    & \textsc{en}   & 316		& 17,479	& 267,359\\
  \midrule
  \textbf{Symbols}      
    & \textsc{en}   & 313		& 1,177		& 283,664\\
  \bottomrule
  \end{tabular}
  \caption{Number of gold, silver and bronze documents per layer and language, as of 13-02-2017. \label{tab:layer_stats}}
\end{table}
\egroup

In addition to adding BoWs in general, we also use annotations to improve the models in a more targeted way, by focusing on annotation conflicts. Annotation conflicts arise when a certain annotation layer for a document has manually checked and marked `gold'.
When the automatic annotation of such a layer changes, e.g., after retraining a model, new annotation errors might be introduced, and these are marked as annotation conflicts. The annotation conflicts are then slated for resolution by an expert annotator.
This has two main benefits: it concentrates human annotation efforts on difficult cases, for which the models' judgements are still in flux, so that the bits of wisdom can steer the model more effectively. In addition, by enforcing conflicts to be re-judged by a human, we have a chance to correct human errors and inconsistencies, and, if necessary, improve the annotation guidelines.

\section{Conclusion}
Our ultimate goal is to provide accurate, language-neutral natural language analysis tools. In the pipeline that we presented in this paper, we have laid the foundation to reach this goal. For every task in the pipeline---tokenization, parsing, semantic tagging, symbolization, semantic interpretation---we have a single component that uses a language-specific model. We proposed new language-neutral tagging schemes to reach this goal (e.g., for tokenization and semantic tagging) and adapted existing formalisms (making CCG more general by introducing lexical categories for empty elements).

Our first results for Dutch show that our method is promising \cite{evang2016}, but we still need to assess how much manual effort is involved in other languages, such as German and Italian. We will also explore the idea of combining CCG parsing with Semantic Role Labelling, following \newcite{lewis:16}, and whether we can derive word senses in a data-driven fashion \cite{Kilgarriff1997} rather than using WordNet.
Furthermore, we will assess whether our cross-lingual projection method yields accurate tools with time and annotation costs lower than would be needed when starting from scratch for a single language.

The annotated data of the PMB is now publicly accessible through a web interface.%
\footnote{\url{http://pmb.let.rug.nl}}
Stable releases will be made available for
download periodically.

\section*{Acknowledgements}
This work was funded by the NWO-VICI grant "Lost in Translation -- Found in Meaning" (288-89-003). The Tesla K40 GPU used for this research was donated by the NVIDIA Corporation. We also wish to thank the two anonymous reviewers for their comments.

\bibliography{eacl2017}
\bibliographystyle{eacl2017}

\end{document}